# Integrating structural and semantic signals in text-attributed graphs with BiGTex

**Azadeh Beiranvand, S. Mehdi Vahidipour***

Faculty of Electrical and Computer Engineering, University of Kashan, Kashan, Iran.

ABSTRACT.

Text-attributed graphs (TAGs) pose unique challenges for representation learning by requiring models to capture both the semantic richness of node-associated texts and the structural dependencies of the graph. While graph neural networks (GNNs) effectively model topological information, they are limited in handling unstructured textual data. Conversely, large language models (LLMs) excel in text understanding but lack awareness of graph structure.

To address this gap, we propose BiGTex, a hybrid architecture built around a novel, modular component: the Graph-Text Fusion Unit. By stacking these units, BiGTex enables tight, bidirectional interaction between textual and structural representations, allowing text to guide structural reasoning and graph topology to refine semantic interpretation within each layer. This stands in contrast to prior sequential or loosely coupled approaches. The model employs parameter-efficient fine-tuning (LoRA), preserving the efficiency of the pre-trained LLM.

Comprehensive experiments on multiple TAG benchmarks demonstrate that BiGTex achieves state-of-the-art performance in node classification and effectively generalizes to link prediction. An ablation study confirms the critical role of bidirectional graph–text interaction in enhancing representational quality.

Overall, this work contributes: (i) a novel hybrid architecture integrating pre-trained language models with GNNs via bidirectional graph–text fusion mechanisms; (ii) a mechanism that injects structural tokens into the LLM and refines representations through learnable graph–text fusion modules; and (iii) empirical evidence showing that BiGTex achieves consistently strong and often superior performance across multiple benchmark datasets compared to state-of-the-art GNN and LLM-enhanced baselines, highlighting its robustness and practical effectiveness in real-world graph-based tasks.

## 1 INTRODUCTION

Many real-world systems can be effectively modelled as graphs, where nodes represent entities and edges capture the relationships between them. In certain types of graphs, nodes are associated with textual information, such as descriptions, labels, or documents. These structures are commonly referred to as Text-Attributed Graphs (TAGs) (Yang et al., 2021). For instance, in Fig.

* Corresponding author.

+ Email addresses: a.Beiranvand@grad.kashanu.ac.ir (Azadeh Beiranvand), vahidipour@kashanu.ac.ir (S. Mehdi Vahidipour).

1, the citation networks (Yang et al., 2016) represent each node as a scientific paper, with its title or abstract serving as textual attributes. Similarly, in social networks (Zeng et al., 2019), nodes usually represent users, each of whom may have a profile containing text-based information such as personal traits, preferences, or interests.

Graph Neural Networks (GNNs) (Hamilton et al., 2017; Kipf & Welling, 2016) have shown strong capabilities in learning high-quality representations from graph-structured data, which are effective in various graph-related tasks such as node classification and link prediction. These models primarily operate based on message passing mechanisms over the graph topology, focusing solely on the structural information. As a result, they inherently overlook any textual attributes associated with the nodes.

On the other hand, Large language models (LLMs) (Vaswani et al., 2017) excel at understanding and representing the semantics of natural language texts. When node attributes are textual, LLMs can extract meaningful features from them. However, LLMs alone are not designed to capture the underlying structure of graphs. Therefore, integrating LLMs with GNNs presents a promising direction, as it enables models to leverage both textual and structural aspects of graph data.

Several approaches have been proposed for integrating LLMs and GNNs. Some methods, such as TAPE (He et al., 2024) and SimTEG (Duan et al., 2023), utilise a language model to encode node texts into dense representations, which are then used as initial node features for a downstream GNN. In these approaches, the LLM and GNN operate independently; the LLM encodes node texts either with or without fine-tuning, and the resulting embeddings are fed into the GNN without considering the graph structure during text encoding.

Other approaches reverse the process of work: a GNN first computes structural embeddings based on graph connectivity, and these are subsequently combined with the original node texts and passed through a language model (e.g., GraphGPT (Tang et al., 2024), Tea-GLM (D. Wang et al., 2024), and GIMLET (Zhao et al., 2023)). While these methods have shown promising results, the language and graph components still function independently, lacking joint optimization or mutual awareness.

More recent approaches attempt to bridge this gap by encouraging alignment between textual and structural representations, often through contrastive learning techniques. For instance, models like GAugLLM (Fang et al., 2024) generate alternative graph views by incorporating LLM-derived embeddings and train the GNN using contrastive objectives to align the original and augmented views.

Despite these advancements, many existing methods either ignore the graph structure during text encoding or fail to update the language model in a graph-aware manner (Ren et al., 2024). This highlights a need for tighter integration between GNNs and LLMs to enable more effective joint representation learning.

Our proposed architecture, BiGTex, leverages the strengths of both language models and graph neural networks by aligning their representations to produce enriched node embeddings for TAGs. The architecture is composed of modular blocks, each integrating a pre-trained language model (PLM) with a GNN layer in a mutually attentive fashion.

Specifically, the output of the GNN layer is used as an additional input token to PLM, placed alongside the textual content of each node. This design encourages the language model to become aware of the graph structure. Simultaneously, the output of PLM is fused with the GNN representation through a fusion block, allowing the GNN to incorporate semantic information from the node texts.

This bidirectional interaction between the two components fosters joint learning of structure-aware and text-sensitive representations. Existing approaches either adopt sequential pipelines (e.g., TAPE (He et al., 2024), SimTeG (Duan et al., 2023)) or loosely coupled joint training (e.g., GLEM (Zhao et al., 2022), ENGINE (Zhu et al., 2024)). However, these strategies lack a systematic mechanism for iterative bidirectional exchange between structure and semantics. BiGTex introduces Graph-Text Fusion Units, modular building blocks that integrate structural embeddings into PLMs as soft prompts, while feeding semantic signals back into GNNs via a type of fusion. This design enables emergent interactions beyond one-way or loosely coupled schemes, positioning BiGTex as a reusable and general framework rather than a task-specific stitching of models. Experimental results demonstrate that our model reliably delivers excellent performance on node classification tasks, whether the pretrained language model is fine-tuned or kept frozen during training. Our main contributions are summarized as follows:

- We introduce BiGTex, a novel hybrid architecture centered on Graph-Text Fusion Units that enable deep, bidirectional integration of pre-trained language models with graph neural networks. Unlike sequential or co-training approaches, our design ensures mutual conditioning of structural and textual representations across all layers.

- Unlike previous methods, our architecture implements a distinct form of dual attention within each block. Specifically, the structural output of the GNN is injected into the language model as an additional token, enabling the language model to attend to graph structure. Meanwhile, the language model's output is refined through a fusion method that works on the GNN embeddings, enabling the graph encoder to incorporate semantic cues from text.

- We conduct extensive experiments across a diverse set of TAG benchmarks. The results demonstrate the effectiveness of our approach, with consistent improvements in performance. Furthermore, we present analyses and ablation studies that demonstrate the importance of each bidirectional component in our architecture.

## 2 PRELIMINARIES

In this section, we provide an overview of the key concepts and background knowledge relevant to our proposed framework. It begins by defining TAGs and tasks in such graphs. Then, it briefly introduces the foundational principles of GNNs and LLMs, followed by a discussion on fine-tuning strategies commonly applied to PLMs in downstream tasks.

### 2.1 Text-Attributed Graph

A text-attributed graph can be formally represented as $G = (V, A, \{S_n\}_{n \in V}, Y)$, where $V$ denotes the set of nodes, $A$ is the adjacency matrix, and $\{S_n\}$ a collection of text sequences associated with the nodes. Each node $n \in V$ is coupled not only with structural features (e.g., neighbourhood connectivity or optional node attributes), but also with a textual description $S_n$, which is a variable-length sequence of tokens of length $L_n$. The label space $Y$ contains node-level annotations for a subset of nodes in $V$, while the remaining nodes are unlabelled (He et al., 2024).

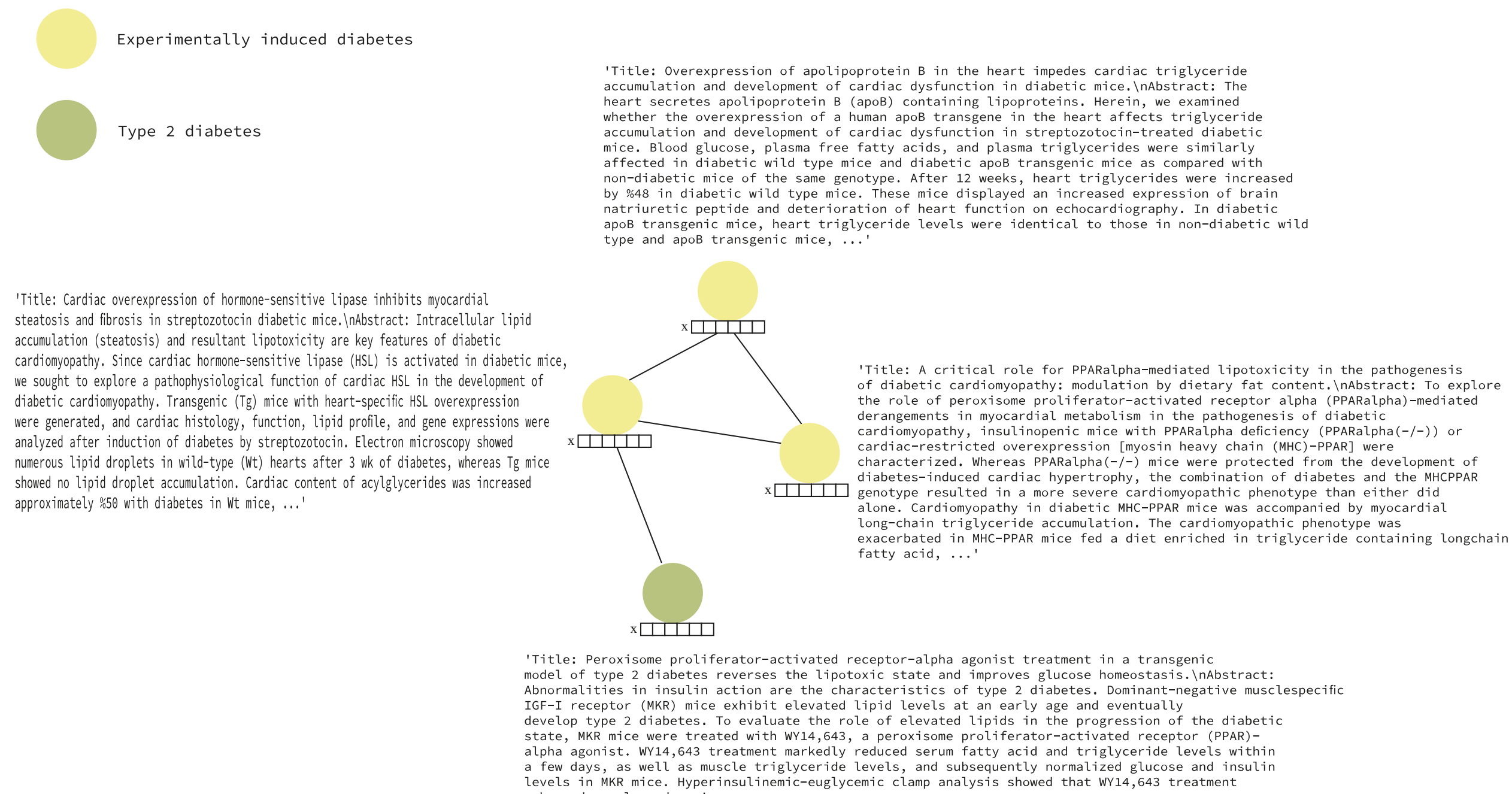

**Fig. 1.** A sample TAG: An example of a graph (citation network) where each node (paper) is associated with a feature vector x and a textual attribute represented as a sequence of tokens (title and abstract).

### 2.2 Task Definitions

In the context of text-attributed graphs, a variety of learning tasks can be formulated based on different supervision levels and application goals. Below, we outline three fundamental tasks commonly studied in this setting.

**Node Classification**: The primary task addressed in this work is node classification, where the goal is to learn a predictive function $f: V \rightarrow Y$. This approach generalizes from labelled nodes to unlabelled ones, using both the graph structure and the associated textual information. Given a partially labelled graph $G$, the objective is to infer the correct labels for the unlabelled nodes based on their position in the graph and the content of their associated texts.

**Link Prediction**: Link prediction focuses on estimating the likelihood of missing or future connections between node pairs in a graph. In a text-attributed setting, this task involves predicting whether an edge should exist between two nodes based on both their structural context (e.g., proximity, common neighbours) and the semantic similarity of their associated text sequences. This task is especially relevant in applications such as recommendation systems, knowledge graph completion, and social network analysis.

**Node Clustering**: Node clustering is an unsupervised task aimed at grouping nodes into clusters such that nodes within the same group exhibit similar structural roles or semantic properties. In text-attributed graphs, this involves jointly leveraging the graph topology and the textual content to identify coherent communities or categories of nodes. Unlike classification, clustering does not rely on labelled data and is often used for exploratory data analysis, visualization, or pre-training of downstream models.

### 2.3 Graph Neural Networks (GNNs)

Graph Neural Networks (Hamilton et al., 2017; Kipf & Welling, 2016) have emerged as powerful tools for learning expressive representations from graph-structured data. They have proven effective in various tasks such as node classification, link prediction, and graph-level inference. The core mechanism behind most GNNs is based on iterative message passing, where each node updates its representation by aggregating information from its local neighbourhood.

Formally, let $\mathcal{N}_n$ denote the set of neighbours of node $n$. At each layer $k$, the representation $\boldsymbol{h}_n^k$ of node $i$ is computed by aggregating the representations of its neighbours from the previous layer $k-1$ and applying an update function:

$$\boldsymbol{h}_n^k = f^k(\boldsymbol{h}_n^{k-1}, AGG(\{\boldsymbol{h}_j^{k-1} : j \in \mathcal{N}_n\})) \in \mathbb{R}^d \tag{1}$$

Here, $AGG$ represents a permutation-invariant aggregation function, such as mean, sum, or max pooling. The function $f$ can be instantiated using a multilayer perceptron, an attention mechanism, or other differentiable transformations. By stacking multiple GNN layers, nodes can incorporate information from multi-hop neighbourhoods, enabling the model to capture richer context within the graph structure.

## 2.4 Large Language Models (LLMs) and Attention Mechanism

Large Language Models are robust deep learning architectures developed for large-scale natural language processing tasks such as text generation, summarization, and classification. These models are typically built on transformer architectures, which avoid sequential computation and instead rely on self-attention mechanisms to model long-range dependencies in language efficiently and in parallel.

A central component in transformers is the attention mechanism, which allows the model to dynamically weigh the importance of different input tokens when computing contextualized representations. This is achieved by transforming each token into three distinct vectors: queries (Q), keys (K), and values (V). The attention weights are calculated using a scaled dot-product function, defined as:

$$Attention(\boldsymbol{Q}, \boldsymbol{K}, \boldsymbol{V}) = softmax\left(\frac{\boldsymbol{QK}}{\sqrt{d_k}}\right)\boldsymbol{V} \tag{2}$$

Here, the dimensionality of the key vectors is used to scale and stabilize gradients during training. This setup enables each token to attend selectively to all others in the input sequence based on learned relevance scores (Vaswani et al., 2017).

To capture different types of relationships simultaneously, the model uses multi-head attention, where multiple independent attention operations are performed in parallel. The outputs of these "heads" are then concatenated and linearly projected to form the final attention output. This design enriches the model's ability to represent diverse semantic and syntactic patterns within text.

LLMs are typically trained in two phases: a pretraining phase and a fine-tuning phase (Bai et al., 2022). During pretraining, the model is exposed to massive corpora using self-supervised objectives such as masked language modelling or causal prediction, allowing it to learn general linguistic patterns without labelled data (Hernández & Amigó, 2021). In the fine-tuning stage, the pretrained model is adapted to specific tasks using smaller, task-specific datasets. Depending on the architecture, this phase may involve full model tuning or lightweight approaches such as adapter layers (Pfeiffer et al., 2020) or LoRA (Hu et al., 2021) (Low-Rank Adaptation).

The synergy between attention mechanisms and large-scale training enables LLMs to extract deep semantic representations, making them highly effective for a broad range of downstream applications—including the one explored in this work.

Continuing from the architectural foundations of large language models, it is important to distinguish between two major classes of transformer-based designs: encoder-only and decoder-only models. Each follows a unique training paradigm and serves different purposes in natural language understanding or generation tasks.

**Encoder-only** models, such as BERT (Devlin et al., 2018), are designed to generate holistic representations of input sequences. These models take an entire sequence as input and process it bidirectionally—each token has access to the whole left and proper context during training. This characteristic allows the model to deeply capture semantic dependencies across the input.

The standard training objective for such models is Masked Language Modelling (MLM), where a random subset of tokens is replaced with a special [MASK] token. The model is then trained to reconstruct the masked tokens using the surrounding unmasked tokens. This approach encourages the network to develop contextualized embeddings grounded in full-sequence comprehension. The loss function for MLM is typically expressed as:

$$\mathcal{L}_{MLM} = -\sum_{i \epsilon \mathcal{M}} \log P(x_i | x_{\setminus \mathcal{M}};\ \theta) \tag{3}$$

where $\setminus \mathcal{M}$ denotes the set of masked positions, $x_{\setminus \mathcal{M}}$ is the unmasked sequence, and $\theta$ represents model parameters.

In contrast, **decoder-only** models, such as GPT (Brown et al., 2020) operate unidirectionally. They are trained in an autoregressive fashion, where each token is generated based only on preceding tokens. The model has no access to future inputs during training or inference, enforcing a strict left-to-right dependency. This makes them particularly well-suited for generative tasks such as open-ended text generation or dialogue modelling.

The pretraining objective for decoder-only models is Causal Language Modelling (CLM), where the model learns to predict the next token in a sequence, given all previous ones. The associated loss is defined as:

$$\mathcal{L}_{CLM} = -\sum_{t=1}^{T} \log P(x_t | x_{<t}; \theta) \tag{4}$$

Here, each token $x_t$ is conditioned solely on the sequence $x_{<t}$, making the model capable of generating coherent text in a step-by-step manner.

The distinction between encoder-only and decoder-only models is foundational: the former are optimized for understanding input context and are widely used in classification and embedding tasks. At the same time, the latter are tailored for sequential prediction and text generation. Choosing between them depends on the nature of the task at hand—whether the goal is to comprehend a complete input or to generate output in a temporally coherent manner.

Training LLMs from scratch is a resource-intensive process, often requiring days of computation across thousands of GPUs and substantial financial investment. As an alternative, fine-tuning pre-trained models has become a widely adopted practice. This approach adapts general-purpose models to specific downstream tasks by continuing training on task-specific data, offering a practical balance between performance and efficiency.

To further reduce the cost of adaptation, Parameter-Efficient Fine-Tuning (PEFT) techniques have been proposed. These methods update only a small subset of parameters while keeping the majority of the pre-trained model frozen, significantly cutting down memory and compute requirements. Among the most effective PEFT methods is LoRA (Hu et al., 2021) (Low-Rank Adaptation), which introduces trainable low-rank matrices into existing weight structures.

Instead of updating the full weight matrix $w_O \in R^{d \times k}$, LoRA approximates the update as a low-rank product:

$$\Delta W = BA \qquad B \in R^{d \times r}, A \in R^{r \times k}, r \ll \min(d,k) \tag{5}$$

This allows the model to adapt to new tasks with minimal parameter updates, while preserving the knowledge encoded in the original model.

In our study, we utilize LoRA as the primary fine-tuning method, enabling efficient adaptation to classification tasks while significantly lowering training costs and resource demands.

While large-scale language models (LLMs) such as GPT, LLaMA, and PaLM have demonstrated remarkable reasoning and generative abilities, they represent only one end of the broader spectrum of pretrained language models (PLMs). PLMs encompass both encoder-based architectures (e.g., BERT, RoBERTa) and generative models (e.g., T5, GPT-series), all built upon large-scale pretraining on textual corpora to capture rich linguistic patterns.

In this work, we employ a compact pretrained language model (specifically, BERT-base) as the textual encoder within our framework. This choice is motivated by two key considerations. First, encoder-based PLMs are highly effective for representation learning tasks, where the goal is to derive semantically meaningful embeddings rather than to generate text. Second, using a smaller PLM substantially reduces computational cost and memory requirements during training, enabling efficient graph–text integration without sacrificing semantic quality. Consequently, our approach retains the contextual understanding and linguistic richness of pretrained models while achieving better scalability and efficiency compared to full-scale LLMs.

## 3 RELATED WORK

The intersection of natural language processing and graph representation learning has given rise to a variety of hybrid approaches for encoding structured data with textual attributes. This section reviews key developments in this evolving area, focusing on how LLMs have been leveraged—either independently or in conjunction with GNNs—to produce expressive node representations.

The existing literatures are categorized into three main directions. First, **Representation Learning with only Large Language Models** explores methods that solely rely on LLMs for representation learning over text-attributed graphs, emphasizing their ability to extract semantic-rich embeddings from node texts (3.1). Next, **Sequential Integration of GNNs and LLMs**, which examines sequential architectures where GNNs and LLMs are applied in a pipeline, with one model generating features that are subsequently refined by the other (3.2). Finally, **Joint Representation Learning with GNN–LLM Integration** discusses more recent efforts that attempt a deeper integration of both paradigms, enabling joint learning and mutual adaptation between structural and textual representations (3.3).

This categorization provides a comprehensive view of the strengths and limitations of current strategies, and positions our proposed framework in the broader context of representation learning on text-attributed graphs.

### 3.1 Representation Learning with only Large Language Models

Recent research explores using LLMs as standalone learners for TAGs, relying solely on textual node attributes while ignoring graph topology. Many studies evaluate off-the-shelf LLMs in zero-shot or frozen settings through prompting and instruction tuning to elicit graph-relevant reasoning (Z. Chen et al., 2024; Guo et al., 2023; Wang et al., 2023). For instance, NLGraph (Wang et al., 2023) benchmarks LLMs on graph reasoning tasks (e.g., connectivity, shortest paths), showing that while careful prompting can yield correct outputs, models remain brittle on complex graphs. GraphGPT (Tang et al., 2024) and related instruction-tuned variants adapt LLMs to better generalize across graph tasks but still lack explicit structural inductive bias.

Some works attempt to implicitly encode structure by linearizing graph neighbourhoods into token sequences using node ordering or serialization techniques (Xypolopoulos et al., 2024), improving contextual understanding yet suffering from token-length constraints that limit scalability. Complementary approaches such as Graph Chain-of-Thought (Jin et al., 2024) enhance reasoning by exposing intermediate graph reasoning steps, although these steps also lack learnable topological modelling.

Several studies fine-tune LLMs for downstream graph tasks (N. Chen et al., 2024; J. Wang et al., 2024; Ye et al., 2024). For instance, GraphWiz (N. Chen et al., 2024) generates task-specific synthetic data and optimizes alignment via Direct Preference Optimization (DPO), while MuseGraph (Tan et al., 2024) serializes neighbourhood information into textual descriptions. Despite these adaptations, scalability and explicit structural reasoning remain significant challenges.

Overall, LLM-only models exhibit strong semantic generalization and context understanding but fail to capture essential graph properties such as local connectivity, homophily, and multi-hop dependencies. To address these limitations, BiGTex bridges LLM semantics and GNN structural awareness through Graph–Text Fusion Units, enabling bidirectional interaction: structural embeddings from the GNN are injected into the LLM as soft prompts. At the same time, semantic outputs feed back into the GNN via cross-attention. This mutual refinement preserves LLM reasoning while embedding learnable topology, yielding coherent representations that unify language and graph modalities.

### 3.2 Sequential Integration of GNNs and LLMs

A second line of work adopts sequential integration between LLMs and GNNs, where both modules operate in a fixed pipeline rather than being jointly optimized. These approaches leverage the semantic expressiveness of LLMs and the structural bias of GNNs, but since each component is trained independently, they often lack true cross-modal adaptation.

In the LLM → GNN direction, node texts are first encoded by an LLM into dense semantic embeddings that serve as node features for graph learning. Examples include TAPE (He et al., 2024) and SimTeG (Duan et al., 2023), where transformer-based language models generate contextual node embeddings later refined by GNNs for node classification and link prediction. Similarly, TextGNN (Zhu et al., 2021) and BertGCN (Lin et al., 2021) use pretrained text encoders like BERT to capture semantic attributes before message passing, while VGCN-BERT (Lu et al., 2020) extends this design by enhancing BERT with graph-derived vocabulary embeddings. These works demonstrate the benefit of incorporating LLM features into GNN pipelines but maintain a unidirectional flow of information from language to structure.

Conversely, the GNN → LLM paradigm feeds graph-informed embeddings into an LLM—typically as concatenated features or soft prompts—to enhance text-grounded reasoning. Representative models include GraphGPT (Tang et al., 2024) and GIMLET (Zhao et al., 2023), which integrate graph-derived knowledge into LLMs for improved zero-shot and instruction-driven understanding. More recent prompting-based frameworks, such as GraphPrompter (Liu et al., 2024) and Graph Neural Prompting (GNP) (Tian et al., 2024), explicitly translate GNN outputs into prompt tokens or embeddings to condition LLM reasoning on relational context.

Despite improved task performance, sequential architectures suffer from unidirectional and lossy information transfer. Since GNN and LLM modules are trained separately, neither component adapts to the other's feedback, and representational inconsistencies remain unresolved. This leads to inefficiencies and limits emergent synergy between structural and semantic signals.

BiGTex addresses these issues by moving beyond fixed pipelines toward iterative bidirectional fusion. Its Graph–Text Fusion Units enable GNN-to-LLM conditioning via soft prompts and LLM-to-GNN feedback via cross-attention within each training iteration. This design preserves the semantic strength of LLM-based pipelines while achieving continuous co-adaptation, producing more coherent and semantically aligned node representations for text-attributed graphs.

**Table 1**
Comparison of joint GNN–LLM integration frameworks and their relation to BiGTex

| Model | Integration Type | Strengths | Limitations | Position vs. BiGTex |
|---|---|---|---|---|
| **GLEM** | Alternating EM | Efficient co-training; label refinement | Asymmetric updates; limited feature exchange | BiGTex enables continuous mutual feedback instead of discrete alternation |
| **ENGINE** | Parallel co-learning | Efficient, modular | Heavy LLM; asymmetric fusion | BiGTex achieves deeper coupling using light LM |
| **GraphAdapter** | Adapter-based | Zero-shot compatibility; plug-and-play | Minimal GNN feedback; no co-adaptation | BiGTex moves beyond unidirectional adaptation by enabling closed-loop, bidirectional feedback (GNN←→LLM) within each fusion unit, fostering co-adaptation. |
| **GreaseLM** | End-to-end | Deep semantic-structural fusion | High compute cost (large LM) | While GreaseLM enables fusion, it often relies on large, frozen LLMs. BiGTex achieves comparable deep fusion with a compact, efficiently tunable LM via bidirectional units, thereby improving parameter efficiency. |
| **DGTL** | Joint contrastive | Strong consistency learning | Complex optimization; limited scalability | BiGTex retains alignment with lower complexity |
| **ConGraT** | Dual-encoder, contrastive pretraining | Strong self-supervised pretraining; transferable embeddings | No fine-grained interaction; relies solely on contrastive loss | BiGTex enables explicit mutual attention and iterative adaptation instead of offline contrastive alignment |
| **Generate** | Joint self-supervised | Captures both textual semantics and structural context; self-supervised; strong generalization | Design complexity in alignment objectives; may require substantial compute for contrastive training | BiGTex achieves similar text-structure alignment via mutual cross-attention using a compact LM and lower training complexity |

### 3.3 Joint Representation Learning with GNN–LLM Integration

Integrating structural and textual information in TAGs remains a key challenge. Unlike sequential pipelines that process modalities independently, recent work focuses on joint optimization frameworks that enable bidirectional exchange between GNNs and LLMs, capturing structure-aware semantics and text-conditioned reasoning within a shared space.

GLEM (Zhao et al., 2022) adopts an Expectation–Maximization scheme where a GNN refines pseudo-labels and an LLM provides semantic supervision, aligning their representations iteratively. ENGINE (Zhu et al., 2024) introduces the G-Ladder architecture, linking an auxiliary

GNN branch to an LLM through learnable adapters for parallel feature exchange and efficient inference. GraphAdapter (Huang et al., 2024) integrates lightweight graph-aware adapters atop frozen LLM layers, improving structural awareness with minimal cost and preserving zero-shot capabilities. GreaseLM (Zhang et al., 2022) extends this idea by enabling bidirectional message passing between transformer layers and GNN modules through cross-modal gating, allowing textual and structural cues to guide each other during reasoning. Further, DGTL (Qin et al., 2023) (Dual Graph–Text Learner) aligns attention maps from both modalities using contrastive learning to enforce local–global consistency, while GRENADE (Li et al., 2023) introduces graph-centric contrastive pretraining for self-supervised representation learning on TAGs, aligning node-level and neighbourhood-level features without labelled data. Similarly, ConGraT (Brannon et al., 2024) aligns GNN and transformer encoders through a cross-modal contrastive objective for joint graph–text embeddings, enabling transferable representations but lacking fine-grained bidirectional adaptation due to its reliance on contrastive loss alone.

Although these methods achieve stronger cross-modal alignment than sequential pipelines, they typically depend on large frozen LLMs (e.g., T5, GPT-J, LLaMA-13B) and additional adapters or co-attention layers, which increase computational cost and limit scalability. Moreover, interactions often remain asymmetric, with stronger influence from LLMs to GNNs, restricting full mutual adaptation. BiGTex addresses these limitations through a lightweight and fully bidirectional fusion strategy. Using a compact transformer encoder coupled with a GNN via mutual cross-attention allows structural embeddings to serve as soft prompts for the LLM. At the same time, semantic outputs feed back into the GNN during message passing. This design enables efficient co-adaptation between modalities, achieving deep semantic–structural alignment without the computational burden of large-scale co-training, and producing coherent and interpretable representations for text-attributed graphs. Table 1 provides a comparative summary of recent GNN–LLM integration frameworks, highlighting how BiGTex differs in both design philosophy and computational efficiency.

As summarized in Table 1, existing joint learning approaches have made significant strides. However, they often exhibit key limitations that our work aims to address: (1) asymmetric or unidirectional interaction (e.g., GraphAdapter primarily conditions the LLM on the graph), where one modality dominates the exchange; (2) reliance on expensive, frozen large LLMs (e.g., GreaseLM) that hinder efficient co-adaptation; or (3) alignment through global contrastive objectives (e.g., DGTL, ConGraT) that may lack fine-grained, instance-level interaction between structure and text. BiGTex is designed to overcome these limitations by introducing stacked Graph-Text Fusion Units that facilitate continuous, bidirectional, and fine-grained feedback within a single, efficient architecture.

## 4 PROPOSED FRAMEWORK: BIGTEX

In this section, we present our proposed framework, which integrates GNNs with LLMs to generate rich and context-aware node representations for TAGs. BiGTex is designed to facilitate deep interaction between textual and structural modalities through a tightly coupled architecture that enables mutual information exchange.

The overall structure of the framework is first outlined, followed by a detailed explanation of its Graph-Text Fusion units (4.1) and training strategy (4.2). A schematic overview of the proposed architecture is illustrated in Fig. 2, where the key modules and the flow of information between the LLM and GNN components are highlighted.

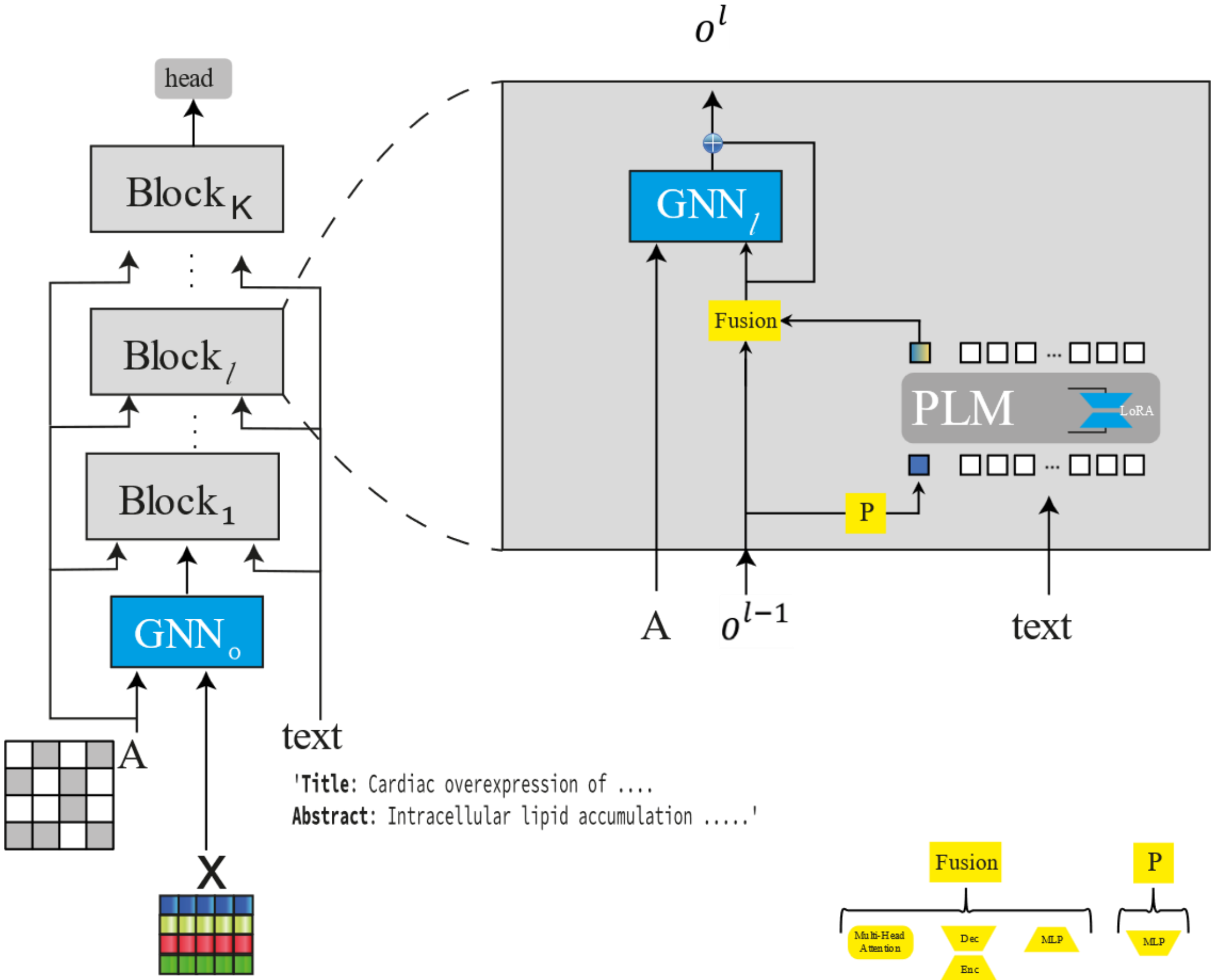

**Fig. 2.** Overview of the proposed architecture (BiGTex). The left side illustrates the overall model structure, where multiple Graph-Text Fusion Units (blocks) are stacked to process input node features and generate final representations. The right side provides a detailed view of a single Fusion Unit, showcasing the internal interactions between the GNN and LLM components. This unit enables structured information exchange between textual and graph representations through tightly coupled fusion operations, facilitating joint representation learning over structural and textual modalities.

Our proposed architecture is composed of a stack of sequential blocks, each referred to as a Graph-Text Fusion Unit, where both a GNN component and a Pretrained Language Model (PLM) module are jointly employed. As illustrated in Fig. 2, each unit receives three inputs:

1) The adjacency matrix of the graph (or the adjacency matrix of the sampled subgraph, A in Fig. 2),
2) The textual descriptions associated with the graph's nodes (text in Fig. 2), and
3) The output representations generated by the preceding Fusion Unit. For the first unit in the stack, the initial node features (typically derived from shallow methods such as Skip-gram, Bag-of-Words, or TF-IDF) are used in place of the previous unit's output (X in Fig. 2).

Within each Graph-Text Fusion Unit, the output vector from the preceding unit is treated as a soft structural token and prepended to the sequence of input tokens corresponding to the node's textual content. A special [SEP] token is inserted between the structural vector and the original text tokens to separate the modalities clearly. This modified input sequence is then passed into the PLM, which computes contextualized token embeddings via its attention mechanism.

Importantly, by prepending the structural embedding to the sequence, the PLM is able to attend to graph-based structural information during its self-attention computations. In this way, the model incorporates graph awareness through a soft-prompting strategy, without modifying the full set of PLM parameters, relying instead on lightweight, parameter-efficient adaptation mechanisms.

### 4.1 Graph-Text Fusion Architecture

We begin by encoding the raw node features with an initial graph neural network layer, denoted as $GNN_0$, which operates solely on the graph structure and adjacency information. This layer produces an initial structure-aware node representation that captures local topological patterns before any interaction with textual information. The purpose of $GNN_0$ is to ground the subsequent multimodal fusion process in a meaningful structural embedding, ensuring that the language model is conditioned on graph-aware signals from the very first fusion stage.

Let $\boldsymbol{o}^{l-1} \in \mathbb{R}^d$ represent the output vector of the node $n$ from the $l-1$-th Fusion Unit. This vector is prepended to the tokenized textual sequence of the node, forming an enriched input to the language model for the next layer. This approach facilitates progressive refinement of node embeddings through layered interaction between textual semantics and structural cues.

$$Prompt_n^{PLM} = [P(\boldsymbol{o}^{l-1}) || [sep] || emb(\{S_n\})] \tag{6}$$

Where $emb(\{S_n\})$ denote the embedding output of the token sequence corresponding to the textual content of the node $n$, and let || represent the concatenation operation. As described earlier, the structural token $o^{l-1}$, obtained from the previous Graph-Text Fusion Unit is concatenated with the textual embeddings to form the input prompt to the language model and P(.) is a projection (a simple MLP) that maps $o^{l-1} \in \mathbb{R}^d$ into $\mathbb{R}^{d_PLM}$ ($d_{PLM}$ *is the PLM's hidden size*). This input enables the PLM to attend not only to the semantic content of the node's text but also to the graph-aware token. Through self-attention, the PLM captures how the structural prompt influences the interpretation of textual tokens. The representation corresponding to the prepended structural position is extracted from the PLM output and treated as a text-conditioned embedding enriched with structural context. In the subsequent step, the interaction flow is reversed: the text-informed representation is fused back into the graph domain. The PLM output is combined with the current graph representation using a lightweight fusion module (e.g., concatenation followed by a projection or bottleneck transformation).

This fused representation is then passed—together with the graph's adjacency structure—into the GNN module of the current Graph-Text Fusion Unit. The GNN then updates the node's representation by aggregating information from its neighbours, now enriched by both structural proximity and contextual textual signals. The output of this GNN layer becomes the final output of the current Fusion Unit and is passed to the next unit in the stack (if any).

#### 4.1.1 Fusion Variants

To integrate textual semantics extracted by the pretrained language model with graph-based structural representations, BiGTex supports multiple fusion strategies. These variants determine how the graph embedding and the text-conditioned embedding are combined within each Graph-Text Fusion Unit. The choice of fusion mechanism is configurable and can be selected to balance model expressiveness and computational complexity.

Let $\boldsymbol{h}_n^{(l)} \in \mathbb{R}^d$ denote the graph-based representation of node $n$ at layer $l$, and let $\boldsymbol{t}_n^{(l)} \in \mathbb{R}^d$ represent the corresponding text-conditioned embedding obtained from the PLM. Three fusion variants are considered: MLP-based fusion, Autoencoder-based fusion, and Cross-Attention-based fusion.

**MLP-based Fusion**

In the MLP fusion variant, graph and text representations are combined through direct concatenation followed by a feed-forward projection. This approach provides a lightweight and efficient mechanism for feature integration while allowing nonlinear interactions between modalities. Formally, the fused representation is computed as:

$$\boldsymbol{z}_n^{(l)} = \phi(\boldsymbol{W}[\boldsymbol{h}_n^{(l)} \,||\, \boldsymbol{t}_n^{(l)}] + \mathbf{b}) \tag{7}$$

where || denotes concatenation, $\boldsymbol{W} \in \mathbb{R}^{2d \times d}$ and $\mathbf{b} \in \mathbb{R}^{d}$ are learnable parameters, and $\phi(\cdot)$ is a nonlinear activation function (e.g., ReLU). This fused vector $\boldsymbol{z}_n^{(l)}$ is then passed to the subsequent GNN layer for neighborhood aggregation.

**Autoencoder-based Fusion (AE)**

The Autoencoder-based fusion introduces a bottleneck structure to encourage compact and informative joint representations. By reconstructing the combined graph–text feature, the autoencoder implicitly enforces alignment between structural and semantic information. First, the concatenated embedding is projected into a latent space:

$$\boldsymbol{z}_n^{(l)} = f_{enc}([\boldsymbol{h}_n^{(l)} \,||\, \boldsymbol{t}_n^{(l)}]) \tag{8}$$

Where $f_{enc}(.)$ denotes the encoder network. The decoder then reconstructs the original concatenated input:

$$\boldsymbol{u}_n^{(l)} = f_{dec}(\boldsymbol{z}_n^{(l)}) \tag{9}$$

The latent representation $\boldsymbol{z}_n^{(l)}$ is used as the fused embedding and forwarded to the GNN. This design encourages the model to preserve complementary information from both modalities while reducing redundancy.

**Cross-Attention-based Fusion**

The cross-attention fusion variant explicitly models directional interactions between graph and text representations. In this setting, the graph embedding serves as the query, while the text-conditioned embedding provides the key and value matrices. The fusion is defined as:

$$\boldsymbol{z}_n^{(l)} = Attention(\boldsymbol{Q} = \boldsymbol{h}_n^{(l)}, \boldsymbol{K} = \boldsymbol{t}_n^{(l)}, \boldsymbol{V} = \boldsymbol{t}_n^{(l)}) \tag{10}$$

where the attention operation follows the standard scaled dot-product formulation based on (2). This mechanism allows the graph representation to selectively attend to semantically relevant textual features, producing a context-aware fused embedding that is subsequently processed by the GNN layer.

Each fusion variant offers a different balance between expressiveness and efficiency. The MLP-based fusion is computationally lightweight, the autoencoder-based fusion promotes compact joint representations, and the cross-attention variant enables fine-grained interaction between modalities. All variants are seamlessly integrated into the proposed framework and can be selected interchangeably without modifying the overall architecture.

### 4.2 Training Strategy and Fine-Tuning

The PLM module in each Graph-Text Fusion Unit can be instantiated with any pretrained language model. If the model is encoder-only—such as BERT—we use the embedding corresponding to the [CLS] token as the output. For decoder-only models like GPT, we instead use the average of the hidden states across all tokens as the representation.

Given that our target task is node classification, we attach a set of lightweight classification heads to the output representations. Specifically, instead of relying on a single shared classifier, we employ a multi-head prediction mechanism, where each class is associated with an independent prediction head implemented as a shallow multi-layer perceptron (MLP). Each head produces a scalar logit corresponding to its target class, and the logits from all heads are concatenated to form the final prediction vector. This design enables class-specific decision boundaries while maintaining a low parameter footprint.

During training, the GNN layers are fully trainable. For the PLM component, we adopt a parameter-efficient fine-tuning (Houlsby et al., 2019) (PEFT) approach using LoRA (Hu et al., 2021) (Low-Rank Adaptation). In this setup, the pretrained weights of the language model remain untouched, while small, trainable low-rank matrices are introduced to enable adaptation to the task. This avoids catastrophic forgetting and significantly reduces the computational cost, allowing the PLM to specialize in the classification task without requiring full fine-tuning.

In the first Fusion Unit, since the initial node features x may not be aligned with the PLM's hidden space ($d_{PLM}$), we apply a linear projection to map the raw input into the correct embedding space. This projection ensures compatibility between the initial input and the PLM-GNN fusion process.

The overall training objective is formulated as a standard cross-entropy loss (Goodfellow et al., 2016) applied to the outputs of the multi-head classifier over labelled nodes:

$$Loss_{cls} = \mathcal{L}_{\theta}(head(\boldsymbol{o}^{K}), \boldsymbol{Y}) \quad (11)$$

where $\boldsymbol{o}^{K}$ is the output representation from the final Fusion Unit, $\boldsymbol{Y}$ denotes the ground truth labels, and $\theta$ includes all trainable parameters: the GNN weights, the LoRA adapters injected into the PLM, the fusion layer, and the parameters of the classification head.

## 5 EXPERIMENTS AND RESULTS

This section presents the experimental setup and evaluates the effectiveness of the proposed model across a range of graph-based tasks.

### 5.1 Experimental Setup

All experiments were conducted under a unified training configuration to ensure fairness and comparability across models. Throughout all experiments, we employ BERT-Base (Devlin et al., 2018) as the default pre-trained encoder unless otherwise specified, language model parameters were frozen during training, and fine-tuning was performed using the LoRA (Low-Rank Adaptation) method with rank=8, allowing for lightweight and parameter-efficient updates. All experiments were executed using PyG 2.5, and computations were performed on a GeForce RTX 4090 GPU. Additional implementation details and dataset-specific experimental settings corresponding to the main results are provided in Appendix C. For full implementation details and code, please refer to our public project page[2].

### 5.2 Datasets

Given that our primary focus is on the node classification task, we evaluate BiGTex on several benchmark datasets widely used in graph learning research: ogbn-Arxiv (Hu et al., 2020), ogbn-Products(subset) (Hu et al., 2020), PubMed (Sen et al., 2008), Ele-Photo (Shchur et al., 2018) and

[2] https://github.com/Azadeh297/BiGTex

Arxiv2023 (He et al., 2024). These datasets offer diverse characteristics in terms of graph size, domain, and textual richness, making them suitable for assessing the generalizability of our approach across different settings. For these datasets, we used raw text data collected in (He et al., 2024). For ogbn-Arxiv, we follow the official data splits provided by the Open Graph Benchmark (OGB) (Hu et al., 2020). For others, we adopt a standard 60/20/20 split for training, validation, and testing, respectively. A summary of their key statistics is provided in Table 2. There are some descriptions of datasets in Appendix A.

**Table 2**
Details of the datasets

| | nodes | edges | Avg. Node Degree | task | Features (dim) |
|---|---|---|---|---|---|
| Cora | 2,708 | 5,429 | 7.79 | 7-class classification | BoW (1433) |
| PubMed | 19,717 | 44,338 | 4.5 | 3-class classification | TF-IDF (500) |
| ogbn-Arxiv | 169,434 | 1,166,243 | 13.7 | 40-class classification | skip-gram (128) |
| Products(subset) | 54,025 | 74,420 | 1.45 | 47-class classification | BoW (100) |
| Arxiv2023 | 46,198 | 78,548 | 3.4 | 40-class classification | word2vec (300) |
| Ele-photo | 48,362 | 500,928 | 36.13 | 12-class classification | TF-IDF (1024) |

### 5.3 Baselines

To evaluate the effectiveness of our proposed model, we compare it against nine baseline methods, grouped into two main categories:

- **GNN-based baselines (MLP&GNNs)**: This category includes classical graph neural network architectures such as GCN (Kipf & Welling, 2016), GAT (Liu & Zhou), and GraphSAGE (Hamilton et al., 2017). These models operate solely on structural and non-textual node features and serve as a reference for evaluating the added benefit of incorporating textual information. Their performance is based on the original node features provided with each dataset (as described in Table 2).

- **LLM-enhanced graph models (GNN&LLM)**: The second group consists of recent state-of-the-art methods that combine large language models with GNNs, either in a sequential or integrated manner. This category includes GAINT (Chien et al., 2022), TAPE (He et al., 2024), SimTeG (Duan et al., 2023), GLEM (Zhao et al., 2022) and ENGINE (Zhu et al., 2024). All of which have demonstrated strong performance on various TAG benchmarks by leveraging both textual and structural signals.

These baselines allow us to assess not only how our method performs relative to classical GNNs but also how it stands against modern hybrid architectures that utilize pretrained language models.

### 5.4 Main results (Node classification task)

The core experimental results are summarized in Table 3. We report the test set accuracies for all models across the selected datasets. All reported values are the mean and standard deviation over five independent runs, each trained for 30 epochs.

As shown in Table 3, the proposed BiGTex framework achieves competitive and, in several cases, superior performance compared to both traditional GNN-based models and recent LLM-enhanced baselines. In particular, BiGTex consistently improves over standalone GNNs (GCN, GAT, and GraphSAGE) across most datasets, highlighting the benefit of integrating textual semantics with graph structural information. Notably, different fusion variants (AE, MLP, and cross-attention) achieve comparable performance, indicating that the overall graph–text interaction paradigm is robust to the specific fusion mechanism used.

**Table 3**
The test accuracy of all models across five text-attributed graph datasets. Each result is reported as the mean ± standard deviation over two runs, using two Graph-Text Fusion Units and 10 training epochs. The best result for each dataset is **bolded**, while the second-best is underlined.

| | | Cora | PubMed | ogbn-Arxiv | Arxiv2023 | Products(subset) | Ele-Photo |
|---|---|---|---|---|---|---|---|
| MLP& GNNs | MLP | 0.7432±0.0275 | 0.8635±0.0032 | 0.5554±0.0011 | 0.6539±0.0039 | 0.5666±0.0010 | 0.6121±0.0011 |
| | GCN | 0.8690±0.0151 | 0.8031±0.0425 | 0.7151±0.0033 | 0.6760±0.0028 | 0.6986±0.0014 | 0.7900±0.0022 |
| | GAT | 0.8573±0.0065 | 0.8850±0.0005 | 0.7192±0.0032 | 0.6906±0.0024 | 0.6975±0.0010 | 0.8035±0.0026 |
| | SAGE | 0.8573±0.0065 | 0.8881±0.0002 | 0.7164±0.0027 | 0.6784±0.0023 | 0.6957±0.0018 | 0.8208±0.0011 |
| GNN&LLM | GIANT | 0.8552±0.0074 | 0.8502±0.0048 | 0.7308±0.0600 | 0.7218±0.2400 | - | 0.8127±0.4100 |
| | TAPE | 0.8842±0.0112 | **0.9618±0.0053** | **0.7750±0.0012** | **0.8423±0.0256** | 0.8234±0.0036 | - |
| | SimTEG | 0.8804±0.0136 | - | 0.7704±0.1300 | 0.7951±0.4800 | - | 0.8307±0.2100 |
| | GLEM | 0.8560±0.0090 | - | 0.7697±0.1900 | 0.7858±0.0900 | - | 0.7610±0.2300 |
| | ENGINE | 0.9148±0.0032 | - | 0.7602±0.2900 | 0.7976±0.1400 | 0.8005±0.4500 | 0.8375±0.0800 |
| BiGTex (AE) | GCN | 0.9244±0.0005 | 0.8865±0.0062 | 0.7290±0.0036 | 0.8112±0.0142 | 0.8712±0.0112 | 0.8279±0.0007 |
| | GAT | 0.9004±0.0018 | 0.8612±0.0037 | 0.7342±0.0017 | 0.8174±0.0114 | 0.8679±0.0023 | 0.8432±0.0014 |
| | SAGE | **0.9306±0.0031** | 0.9336±0.0014 | 0.7367±0.0117 | 0.8228±0.0102 | 0.8598±0.0075 | **0.8536±0.0154** |
| BiGTex (MLP) | GCN | 0.9143±0.0054 | 0.9096±0.0035 | 0.7200±0.0216 | 0.8112±0.0075 | 0.8642±0.0142 | 0.8324±0.0132 |
| | GAT | 0.9115±0.0009 | 0.9112±0.0047 | 0.7352±0.0147 | 0.8205±0.0034 | 0.8721±0.0214 | 0.8379±0.0211 |
| | SAGE | 0.9251±0.0014 | 0.9323±0.0115 | 0.7310±0.0124 | 0.8212±0.0121 | 0.8702±0.0078 | 0.8359±0.0043 |
| BiGTex (cross-att) | GCN | 0.9142±0.0018 | 0.8990±0.0025 | 0.7150±0.0140 | 0.8075±0.0116 | 0.8697±0.0162 | 0.8352±0.0141 |
| | GAT | 0.9201±0.0024 | 0.9054±0.0045 | 0.7275±0.0088 | 0.8105±0.0141 | 0.8624±0.0241 | 0.8341±0.0210 |
| | SAGE | 0.9262±0.0062 | 0.9126±0.0021 | 0.7326±0.0241 | 0.8230±0.0117 | **0.8759±0.0142** | 0.8488±0.0075 |

In addition, BiGTex exhibits strong performance on citation networks with moderate textual content, such as PubMed and Cora. On Cora, BiGTex consistently outperforms both pure GNN models and MLP-based baselines, suggesting that even relatively short node texts can effectively complement graph structure when jointly modelled. This result is particularly important, as Cora represents a widely used benchmark where node features are often limited and noisy. It also demonstrates strong generalization to sparser-text datasets like Ele-Photo. In this dataset, node texts often consist of short or noisy product descriptions—sometimes limited to a few keywords or even a single token (e.g., "good")—which provide little semantic context on their own. Despite this sparsity, BiGTex achieves competitive performance, indicating that its fusion mechanism effectively compensates for weak textual signals by leveraging structural cues from neighboring nodes.

Overall, these results demonstrate that BiGTex can effectively leverage textual attributes to enhance node representations, while still benefiting from structural neighbourhood aggregation. The consistent improvements across datasets with varying graph sizes and text richness indicate that the proposed framework is well-suited for a broad range of text-attributed graph learning scenarios.

### 5.5 Visualizing Embedding Quality with t-SNE

To qualitatively evaluate the discriminative power of the learned embeddings, we use t-SNE (Van der Maaten & Hinton, 2008) to project the high-dimensional node representations into a 2D space. For each dataset, we randomly sample 100 nodes per class and visualize them as individual points, colour-coded by class.

As illustrated in Fig. 3 We compare three types of representations:

- **X-orig**: the original input features provided by the dataset,
- **X-SimTeG**: node embeddings obtained from the SimTeG (Duan et al., 2023) model,
- **X-TAPE**: node embeddings obtained from the TAPE (He et al., 2024) model,
- **X-BiGTex**: node representations generated by our proposed method.

The t-SNE plots are shown for two datasets: ogbn-Arxiv and Arxiv2023. In both cases, the embeddings produced by our model exhibit clearer class separation compared to those of the others. This observation suggests that our model effectively integrates both textual and structural information, resulting in more informative and separable representations in the latent space.

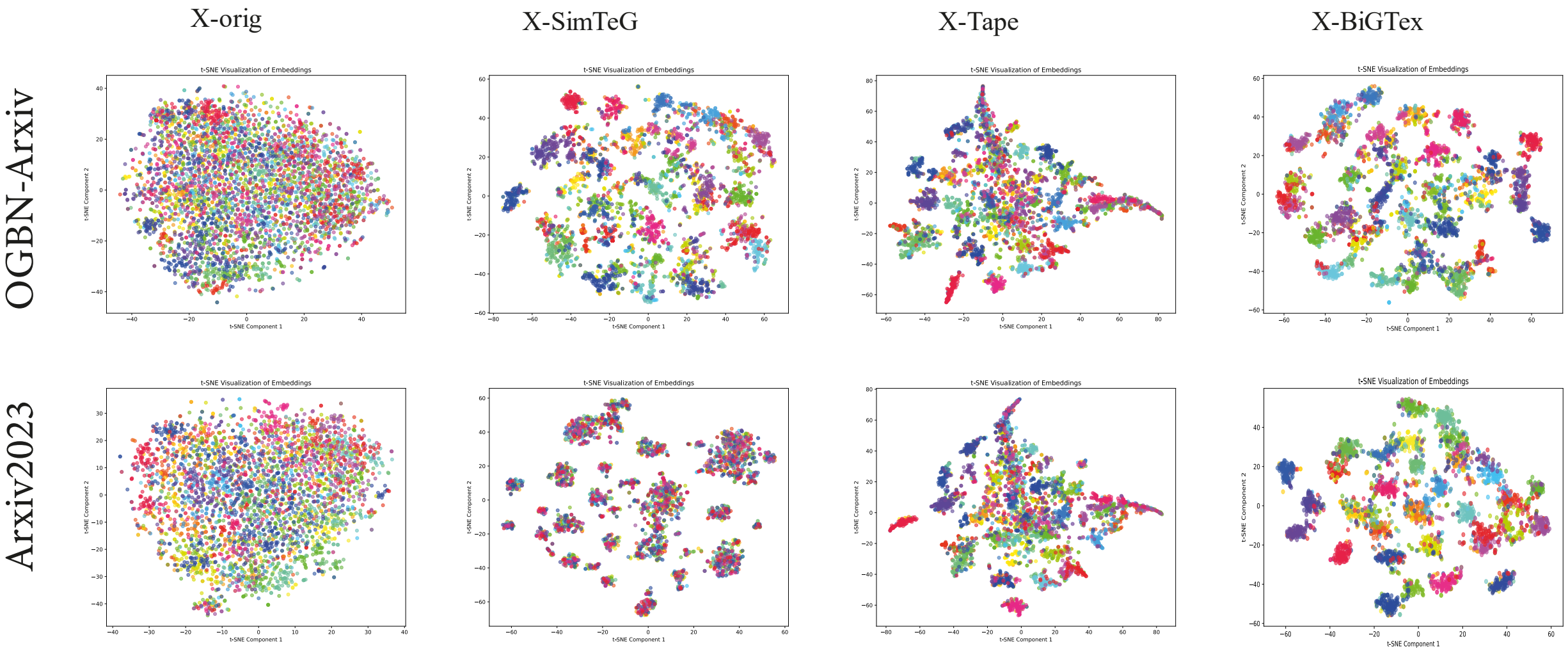

**Fig. 3.** t-SNE visualization of node representations for ogbn-Arxiv and Arxiv2023 datasets. Each plot shows a 2D projection of 100 randomly sampled nodes per class. From left to right: original input features (X-orig), embeddings generated by SimTeG (Duan et al., 2023) (X-SimTeG), embeddings generated by TAPE (He et al., 2024) (X-TAPE) and embeddings generated by our proposed model (X-BiGTex). Compared to the baselines, our method produces more distinct and well-separated clusters, indicating stronger class-level discrimination in the learned representations.

### 5.6 Link Prediction Task

To assess the versatility of the learned representations, we further evaluate our model on a link prediction task. Specifically, we reuse the node embeddings generated by our model—originally trained for node classification—and apply them to predict the existence of edges between node pairs. The performance is measured using the Area Under the ROC Curve (AUC) metric. As shown in Table 4, we compare the results of our method with those of SimTeG and TAPE, strong baselines that also integrate language models with GNNs. Although our embeddings were obtained through supervision on a node classification task, without any task-specific fine-tuning for link prediction, they achieve competitive and, in several cases, superior AUC scores compared to the baseline methods. In particular, BiGTex consistently outperforms TAPE across all evaluated datasets, and demonstrates strong competitiveness with SimTeG—achieving higher AUC on Arxiv2023, while remaining comparable on ogbn-Arxiv.

**Table 4**
AUC scores for link prediction.

| | X-orig | X-SimTeG | X-TAPE | X-BiGTex |
|---|---|---|---|---|
| ogbn-Arxiv | 0.7226 | **0.9704** | 0.8831 | <u>0.9527</u> |
| Arxiv2023 | 0.8033 | 0.8546 | <u>0.8611</u> | **0.9081** |

Moreover, all text-enhanced methods substantially outperform the original graph-only baseline (X-orig), highlighting the importance of incorporating textual information into graph representations. These results underscore the generalizability of our model: the dual-modality embeddings learned through bidirectional interaction between text and structure are not only effective for node classification, but also transferable to structurally different downstream tasks such as link prediction. This suggests that the integration of bidirectional attention between textual and structural modalities fosters representations that are robust, flexible, and broadly applicable across multiple graph-based learning scenarios.

### 5.7 Ablation Study

To investigate the individual impact of key components in our architecture, we conduct an ablation study by selectively removing each element and evaluating the resulting model performance. The results are summarized in Table 5. Our full model, referred to as BiGTex, consistently achieves the best or near-best performance across all datasets, confirming the effectiveness of the complete architecture. When LoRA-based fine-tuning is disabled (w/o LoRA), we observe a noticeable drop in performance across all datasets. This degradation is more pronounced on larger and more complex graphs such as ogbn-Arxiv and Products, indicating that LoRA plays an important role in adapting the pretrained language model to graph-based learning while maintaining parameter efficiency. Removing the soft prompt mechanism (w/o soft prompt: the structural token $P(\boldsymbol{o}^{l-1})$ is removed from the input to the PLM (6), disabling the GNN→PLM conditioning) also leads to a consistent performance decline, demonstrating the importance of injecting structural information directly into the language model's input space. While the drop is generally smaller than that observed when removing LoRA alone, the results suggest that soft prompting contributes meaningfully to aligning textual representations with graph structure. Notably, when both LoRA and soft prompting are omitted (w/o soft prompt & LoRA), the model performance degrades substantially across all datasets, with particularly large drops on Products and Arxiv2023. This behavior highlights the complementary nature of these two components and confirms that their combination is critical for effective graph–text representation learning.

Finally, we examine the effect of removing the initial graph neural network layer (w/o $GNN_0$). The performance decrease across all datasets—most notably on Arxiv2023 and ogbn-Arxiv—confirms that $GNN_0$ plays a crucial role in extracting localized structural features before fusion with textual representations. While the drop is generally less severe than when removing LoRA or both LoRA and soft prompting, it underscores that early graph-based aggregation provides a foundational signal that benefits subsequent text–graph alignment.

Overall, the ablation results underscore that neither LoRA nor soft prompting alone is sufficient to fully realize the benefits of the proposed framework; instead, their joint use enables robust and consistent performance gains across diverse text-attributed graph datasets.

**Table 5**
Ablation study results showing the test accuracy of the full model (BiGTex) and its reduced variants across five datasets. Removing either LoRA or soft prompt leads to noticeable performance drops, while omitting both causes severe degradation.

| | Cora | PubMed | ogbn-Arxiv | Arxiv2023 | Products(subset) | Ele-photo |
|---|---|---|---|---|---|---|
| BiGTex | **0.9306** | **0.9336** | **0.7367** | **0.8230** | **0.8759** | **0.8536** |
| w/o LoRA | 0.8512 | 0.8654 | 0.6841 | 0.7642 | 0.8102 | 0.7941 |
| w/o soft prompt | 0.9125 | 0.9121 | 0.7201 | 0.8163 | 0.8624 | 0.8410 |
| w/o soft prompt & LoRA | 0.8214 | 0.8595 | 0.6547 | 0.6024 | 0.5962 | 0.7892 |
| w/o $GNN_0$ | 0.9124 | 0.9142 | 0.7212 | 0.7957 | 0.8612 | 0.8431 |

### 5.8 Post-processing with Correct & Smooth on ogbn-Arxiv

To further investigate the potential of post-processing techniques to improve node classification performance, we conduct an additional experiment on the ogbn-Arxiv dataset using the Correct and Smooth (C&S) method (Huang et al., 2021). In this experiment, we first select the best-performing BiGTex configuration on ogbn-Arxiv based on the main results reported in Section 5.4. The node-level prediction logits produced by this model are then refined using the Correct and Smooth procedure.

We focus this analysis on ogbn-Arxiv for two main reasons. First, ogbn-Arxiv is a large-scale, citation-based graph with strong homophily, for which label-propagation–based techniques such as C&S are particularly effective. Second, due to its dense structural connectivity and rich citation patterns, ogbn-Arxiv provides a suitable testbed for evaluating whether graph-aware post-processing can further enhance the representations learned by our graph–text fusion architecture.

The motivation behind applying C&S in this setting is to examine whether the representations learned by BiGTex already encode sufficient local and global structural consistency to benefit from label smoothing and error correction. Since BiGTex jointly incorporates textual semantics and graph topology during training, we hypothesize that its prediction errors are more likely to be locally coherent rather than random. Under this assumption, C&S can effectively propagate reliable label information across neighbouring nodes while correcting isolated misclassifications.

**Table 6**
Impact of Correct and Smooth (C&S) post-processing on ogbn-Arxiv results

| | Validation Accuracy | Test Accuracy | Rank in OGB Leaderboard[3] |
|---|---|---|---|
| BiGTex | 0.7461±0.0216 | 0.7367±0.0117 | 37 |
| BiGTex+C&S | 0.7470±0.0046 | 0.7424±0.0033 | 22 |

As shown in Table 6, applying Correct and Smooth on top of BiGTex leads to a noticeable improvement in classification accuracy on ogbn-Arxiv. This result suggests that the node embeddings produced by BiGTex are well aligned with the underlying graph structure, enabling post hoc propagation methods to further refine the decision boundaries. Importantly, this improvement is achieved without additional model training, highlighting the complementary nature of BiGTex and classical graph-based post-processing techniques.

Overall, this experiment demonstrates that BiGTex not only performs competitively as a standalone model but also serves as a strong foundation for downstream refinement methods. The combination of deep graph–text fusion during training and lightweight structural post-processing at inference time offers a flexible and effective strategy for large-scale text-attributed graph learning.

[3] Leaderboards for Node Property Prediction | Open Graph Benchmark

### 5.9 Effect of Stacked Graph-Text Fusion Units on BiGTex Performance

We systematically evaluate the effect of architectural depth in BiGTex by varying the number of stacked Graph-Text Fusion Units from 1 to 5. Each unit serves as a self-contained module for hierarchical integration of structural and textual information. Fig. 4 presents the performance trends across five benchmark datasets: Cora, Products, Arxiv2023, ogbn-Arxiv, Ele-Photo, and PubMed.

Empirically, we observe a substantial performance gain when increasing the number of stacked units from 1 to 2 across all datasets. This sharp improvement indicates that stacking multiple Fusion Units significantly enhances BiGTex's ability to capture complex interactions between graph topology and node-level textual semantics. Beyond two units, the performance gains become more moderate and exhibit dataset-dependent behavior. While certain datasets, such as PubMed and ogbn-Arxiv, benefit from deeper architectures with three stacked units, others show marginal improvements or slight fluctuations as the depth increases. This suggests that although deeper stacking can further refine representations, most of the expressive power is already achieved with two Fusion Units, offering a favorable balance between performance and model complexity.

Overall, these results demonstrate that a small number of stacked Graph-Text Fusion Units is sufficient to effectively integrate structural and textual information, while deeper architectures yield diminishing returns depending on the dataset characteristics.

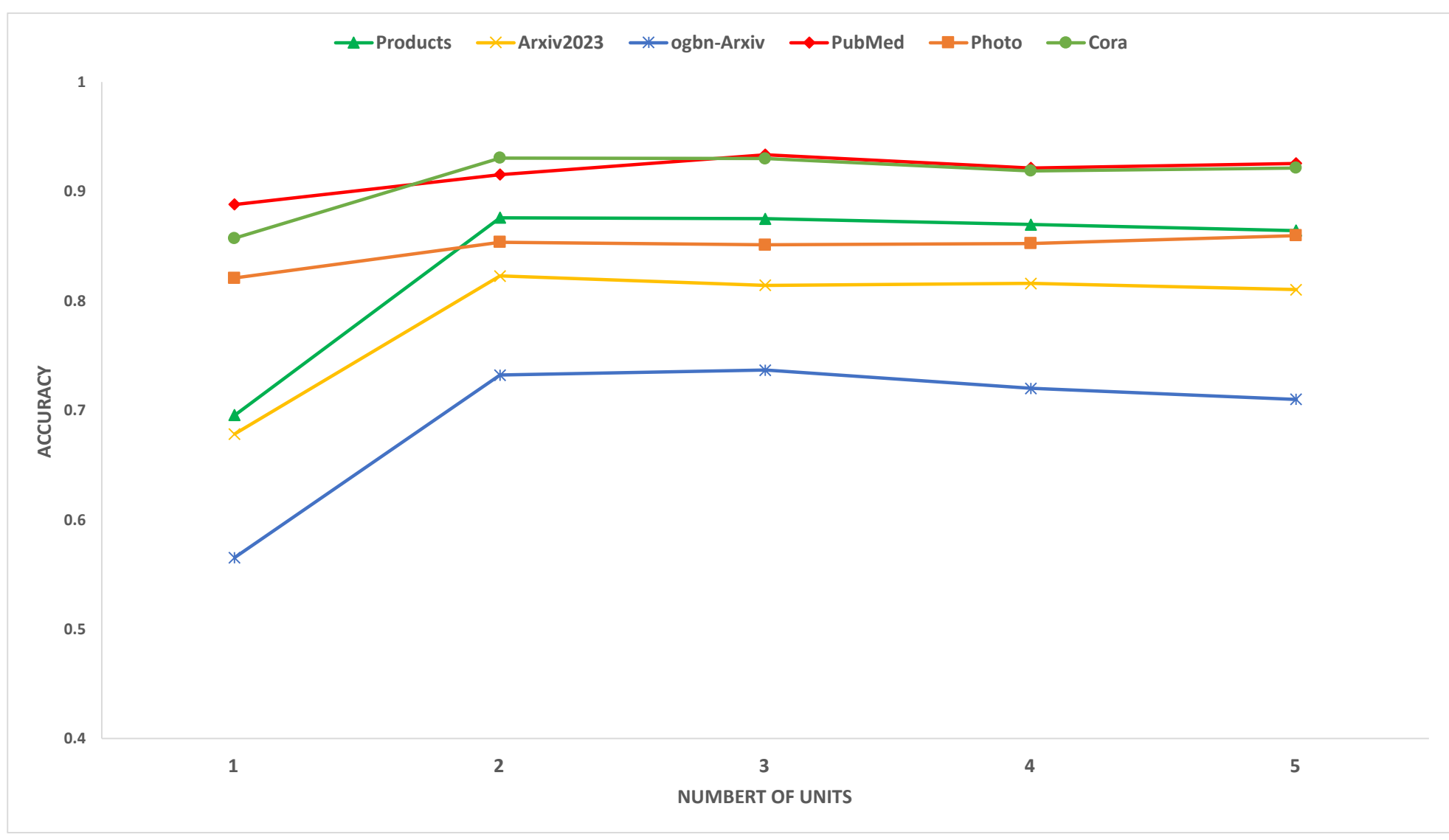

**Fig. 4**. Performance variation of BiGTex with respect to the number of stacked Graph-Text Fusion Units on Cora, Products, Arxiv2023, ogbn-Arxiv, and PubMed datasets. The most significant performance improvement is observed when increasing the depth from one to two units.

Theoretically, this behavior aligns with the balance between expressiveness and overfitting. While stacking more Fusion Units increases model capacity and enables the capture of higher-order interactions, it also introduces challenges such as overfitting, optimization instability, and diminishing returns — especially in moderately sized graphs where textual features already embed rich semantics. These results suggest that for text-attributed graphs, BiGTex benefits most from a moderate depth of two stacked Graph-Text Fusion Units, achieving strong representational power without unnecessary architectural complexity. This finding highlights the importance of architectural parsimony when designing graph-text fusion models.

## 6 CONCLUSIONS, LIMITATIONS, AND FUTURE STUDIES

In this paper, we introduced BiGTex, a hybrid architecture that combines the strengths of graph neural networks and large language models through deep, bidirectional interaction. By integrating a soft structural prompt into the PLM input and using fusion mechanisms to propagate textual cues back into the GNN, our model enables rich and task-relevant node representations to emerge. The key contribution of BiGTex lies not merely in combining GNNs and LLMs, but in introducing Graph-Text Fusion Units as a reusable architectural primitive. By unifying soft prompting (for injecting structural context into LLMs) and cross-attention (for integrating semantic signals into GNNs), BiGTex fosters emergent interactions that sequential or loosely coupled strategies cannot capture. This modular design allows BiGTex to generalize across tasks and backbones, positioning it as a step toward a foundation-style framework for text-attributed graphs. Experimental results across a variety of TAG datasets demonstrate that BiGTex not only surpasses both traditional GNNs and existing LLM-enhanced baselines but also produces embeddings that transfer well to other tasks, such as link prediction, even when trained solely under node-level supervision. Our ablation analysis confirms the complementary roles of structural prompting and parameter-efficient fine-tuning.

**Theoretical Underpinning and Design Rationale**. The efficacy of BiGTex stems from its principled, bidirectional design. First, the bidirectional flow of information addresses the semantic-structural gap by creating a closed-loop refinement process: the GNN-to-PLM path (via soft prompts) grounds language generation in local graph structure, reducing hallucination and increasing contextual relevance, while the PLM-to-GNN path (via fusion modules) enriches structural messages with disambiguating semantic cues. This iterative process can be viewed as progressively maximizing the cross-modal mutual information between structural and textual representations, leading to more robust and aligned embeddings. Second, with respect to scalability and overfitting, stacking multiple fusion units is designed for progressive refinement rather than merely increasing the number of parameters. The risk of over-smoothing or redundancy is mitigated by (a) residual connections that preserve information from earlier layers, and (b) the adaptive gating or attention mechanisms within each fusion unit, which learn to focus on novel, complementary cross-modal signals at each step. This design encourages each unit to refine the representation meaningfully, preventing degeneracy.

While BiGTex achieves strong and consistent results, several aspects warrant further exploration.

**Computational Efficiency and Benchmarking**: Although BiGTex employs a compact LM and parameter-efficient tuning (LoRA), its computational demands can grow with the number of fusion units, very large graphs, or long texts. While our work provides a theoretical complexity analysis and reports parameter counts, a direct, controlled empirical comparison of runtime and memory usage against all baselines on identical hardware remains future work. Such a unified benchmark, though highly valuable for practical adoption, requires careful replication under the same conditions and is an important avenue for community-wide evaluation. Future extensions could further improve scalability through lightweight fusion modules, sparse attention, or progressive message-passing schemes.

**Negative Sampling for Link Prediction**: In this work, negative edges are sampled randomly. While standard, this can introduce false negatives among semantically similar nodes. Future work could incorporate semantic-aware or hard-negative sampling to enhance the robustness and interpretability of link prediction, particularly in dense or semantically overlapping graphs.

**Handling Sparse or Noisy Text**: BiGTex assumes reasonably informative node text. Real-world graphs often have sparse, noisy, or minimal text (e.g., short descriptions). Encouragingly, BiGTex's bidirectional mechanism already compensates for weak text by leveraging structural context. To further strengthen this, future research could explore graph–language augmentation techniques like retrieval-augmented generation or LLM-driven synthetic node generation (Yu et al., 2025), combined with BiGTex's fusion units, to create a more robust framework for low-text scenarios.

**Extension to Heterogeneous and Dynamic Graphs**: Our current design focuses on static, homogeneous graphs. Extending BiGTex to heterogeneous or dynamic graphs (with evolving node/edge types) is a promising direction. This would likely require adaptive fusion units that model temporal dependencies and type-specific message passing while preserving efficient bidirectional alignment.

## Ethical approval and consent to participate

This study does not involve human participants, animal subjects, or any personally identifiable data. All datasets used are publicly available and cited correctly. The authors confirm adherence to research integrity and ethical standards in data handling, experimentation, and reporting.

## Funding

This research received no external funding. The work was conducted using publicly available datasets and open-source software tools.

## Declaration of competing interest

The authors declare that they have no known competing financial interests or personal relationships that could have influenced the work reported in this paper.

## Declaration of generative AI and AI-assisted technologies in the manuscript preparation process

During the preparation of this work, the author(s) used AI-assisted tools (e.g., grammar and spell-checkers, literature search assistants) solely to improve language clarity and to support literature exploration. After using these tools, the author(s) thoroughly reviewed and edited all content, and take(s) full responsibility for the scientific accuracy, analysis, interpretation, and conclusions presented in the published article.

## APPENDIX A: DATASET DESCRIPTION

**Cora (**Sen et al., 2008**)** is a citation network widely used for benchmarking graph-based learning methods. Each node represents a scientific publication in the field of machine learning, while edges denote citation relationships between papers. The graph contains 2,708 nodes and 5,429 edges. The classification task assigns each publication to one of seven predefined research categories, such as neural networks, probabilistic methods, or reinforcement learning. Accuracy is adopted as the primary evaluation metric for this dataset.

**PubMed** (Sen et al., 2008) is a biomedical citation network constructed from research articles indexed in the PubMed database. Each node represents a scientific publication related to diabetes, and directed edges indicate citation relationships between articles. The graph contains 19,717 nodes and 44,338 edges. The classification task involves assigning each article to one of three medical categories based on its subject area. Accuracy is used as the primary evaluation metric for this dataset.

**ogbn-Arxiv** is a large-scale citation network sourced from the Open Graph Benchmark (Hu et al., 2020) (OGB). Each node represents a scientific paper from the computer science domain posted on arXiv, and edges correspond to citation links directed from newer papers to older ones. The graph includes approximately 169,343 nodes and 1.1 million edges. The task is to predict the subject category of each paper among 40 classes. The standard data split provided by OGB is used, and node classification accuracy serves as the evaluation metric.

**ogbn-Products (subset) and ogbn-Products** (Hu et al., 2020) is an undirected product co-purchasing network derived from Amazon. In this graph, each node represents a product, and an edge connects two products if they are frequently bought together. The classification task involves predicting the product category among 47 distinct classes. The dataset follows the official split provided by the Open Graph Benchmark, and accuracy is used as the primary evaluation measure.

**Arxiv2023** (He et al., 2024) is a curated citation network derived from a subset of arXiv papers published in 2023. Each node corresponds to a scientific article, and directed edges reflect citation links among them. The dataset is designed specifically for evaluating text-attributed graph models, where each node is associated with rich textual metadata such as the title and abstract. The task is to classify nodes into a predefined set of subject areas, with performance measured using accuracy.

**Ele-Photo** (Shchur et al., 2018) is a product co-purchase network derived from the Amazon Electronics and Photo categories. Each node corresponds to a product listed on Amazon, and edges represent co-purchasing relationships between items frequently bought together. The textual feature of each node is the product description or user review summary, which varies from detailed sentences to short keywords (e.g., "good" or "recommended"). The task is to classify each product into one of several high-level categories, such as Electronics or Photo equipment. Despite the sparsity and noise in textual attributes, this dataset provides a realistic benchmark for evaluating models like BiGTex that must integrate weak semantic information with structural cues. The dataset contains 7,650 nodes and 119,081 edges for Amazon Photo, and 13,381 nodes with 245,778 edges for Amazon Electronics. Accuracy is used as the primary evaluation metric.

## APPENDIX C: DATASET-SPECIFIC EXPERIMENTAL SETTINGS

This appendix provides detailed experimental configurations for each dataset used in our evaluation. For clarity and reproducibility, we report the dataset-specific architectural choices, language model backbones, and optimization hyperparameters used for the main results. The complete configuration for each dataset is summarized in Table 7

**Table 7**
Experimental configurations for all datasets

| Dataset | PLM | #Layers | Hidden Size | LR | Weight Decay | #Epoch | Batch size | #Parameters |
|---|---|---|---|---|---|---|---|---|
| Cora | *BERT-BASED* | 2 | 768 | 1e-4 | 5e-4 | 30 | 64 | 5,159,751 |
| PubMed | *SciBERT* | 3 | 768 | 1e-4 | 5e-4 | 30 | 64 | 6,023,619 |
| ogbn-Arxiv | *SciBERT* | 3 | 768 | 1e-4 | 5e-4 | 30 | 64 | 7,799,400 |
| Products(subset) | *BERT-BASED* | 2 | 768 | 1e-4 | 5e-4 | 30 | 64 | 6,713,962 |
| Arxiv2023 | *BERT-BASED* | 2 | 768 | 1e-4 | 5e-4 | 30 | 64 | 6,542,438 |
| Ele-photo | *BERT-BASED* | 2 | 768 | 1e-4 | 5e-4 | 30 | 64 | 5,321,036 |